\documentclass[letterpaper, 10 pt, conference]{ieeeconf}  

\IEEEoverridecommandlockouts                              

\usepackage{graphics} 
\usepackage{epsfig} 
\usepackage{mathptmx} 
\usepackage{times} 
\usepackage{amsmath} 
\usepackage{amssymb}  
\usepackage{subcaption}
\usepackage{cite}  
\usepackage[]{algorithm} 
\usepackage{algpseudocode} 
\usepackage{float}  
\usepackage{graphicx}
\usepackage{placeins}
\usepackage{balance}
\usepackage{xcolor}
\title{\LARGE \bf
GANVO:	Unsupervised Deep Monocular Visual Odometry and Depth Estimation with Generative Adversarial Networks
}

\author{Yasin Almalioglu$^{1}$, Muhamad Risqi U. Saputra$^{1}$, Pedro P. B. de Gusmão$^{1}$, Andrew Markham$^{1}$, and  Niki Trigoni$^{1}$
\thanks{$^{1}$Yasin Almalioglu, Muhamad Risqi U. Saputra, Pedro P. B. de Gusmão, Andrew Markham, and Niki Trigoni are with the Computer Science Department, The University of Oxford, UK
        {\tt\small \{yasin.almalioglu, muhamad.saputra, pedro.gusmao, andrew.markham, niki.trigoni\}@cs.ox.ac.uk}}%
}

\begin{document}

\maketitle
\thispagestyle{empty}
\pagestyle{empty}

\begin{abstract}
In the last decade, supervised deep learning approaches have been extensively employed in visual odometry (VO) applications, which is not feasible in environments where labelled data is not abundant. On the other hand, unsupervised deep learning approaches for localization and mapping in unknown environments from unlabelled data have received comparatively less attention in VO research. In this study, we propose a generative unsupervised learning framework that predicts 6-DoF pose camera motion and monocular depth map of the scene from unlabelled RGB image sequences, using deep convolutional Generative Adversarial Networks (GANs). We create a supervisory signal by warping view sequences and assigning the re-projection minimization to the objective loss function that is adopted in multi-view pose estimation and single-view depth generation network. Detailed quantitative and qualitative evaluations of the proposed framework on the KITTI \cite{menze2015object} and Cityscapes \cite{cordts2016cityscapes} datasets show that the proposed method outperforms both existing traditional and unsupervised deep VO methods providing better results for both pose estimation and depth recovery.
\end{abstract}

\section{INTRODUCTION}

Visual odometry (VO) and depth recovery are essential modules of simultaneous localization and mapping (SLAM) applications. In the last few decades, VO  systems have attracted a substantial amount of attention, enabling robust localization and accurate depth map reconstruction. Monocular VO is confronted with numerous challenges such as large scale drift, the need for hand-crafted mathematical features and strict parameter tuning \cite{engel2014lsd, mur2017orb}. Supervised deep learning based VO and depth recovery techniques have showed good performance in challenging environments and succesfuly alleviated issues such as scale drift, need for feature extraction and parameter finetuning \cite{wang2017deepvo, clark2017vinet, turan2018deep, turan2017deep}. VO as a regression problem in supervised deep learning exploits the capability of convolutional neural network (CNN) and recurrent neural network (RNN) to estimate camera motion, to calculate optical flow, and to extract efficient feature representations from raw RGB input \cite{wang2017deepvo, clark2017vinet, muller2017flowdometry, turan2018deep}. In recent years, unsupervised deep learning approaches have achieved remarkable results in various domains eliminating the need for labelled data \cite{artetxe2017unsupervised, turan2018unsupervised}. 

\begin{figure}[t]
\centering
\includegraphics[width=\columnwidth]{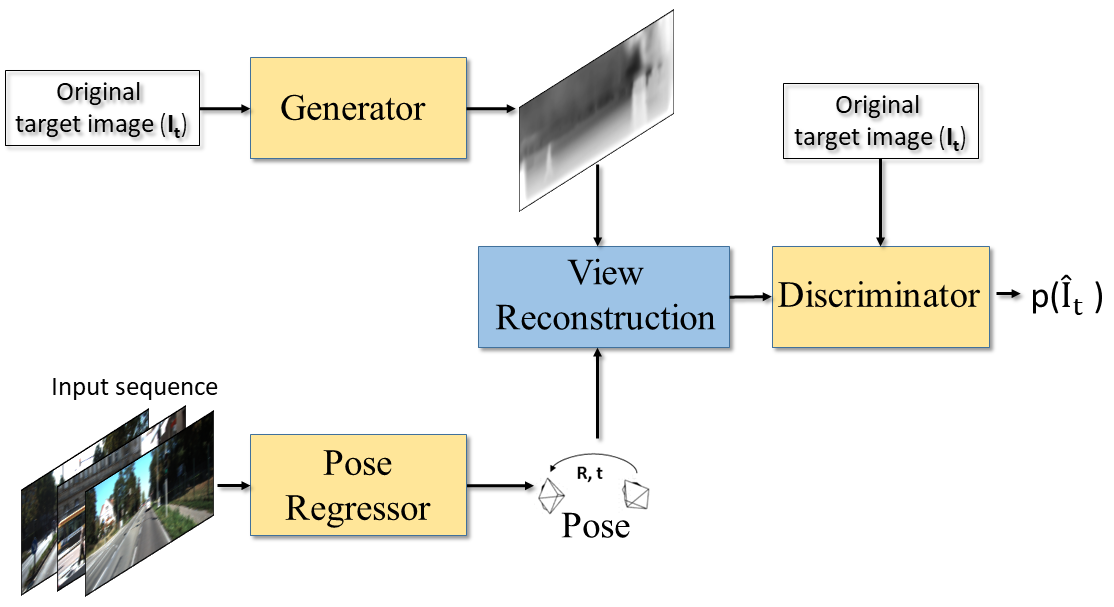}
\caption{Architecture overview. The unsupervised deep learning approach consists of depth generation, multi-view pose estimation, view reconstruction, and target discrimination modules. Unlabelled image sequences from different temporal points are given to the networks to establish a supervision signal. The networks estimate relative translation and rotation between consecutive frames from different perspectives parametrized as 6-DoF motion, and depth image as a disparity map for a given view. } 
\label{fig:model_overview}
\end{figure}

Years of research in visual SLAM have been inspired by human navigation that easily locates obstacles even in unknown environments. A neuroscientific insight is that human brain saves a structural perception of the world, which makes it capable of real and imaginary scene reconstruction through vast previous experiences \cite{brown2016prospective,wamsley2010brief}. In this study, we propose a novel real-time localization and generative depth estimation approach that mimics the remarkable ego-motion estimation and re-constructive scene representation capabilities of human beings by training an unsupervised deep neural network. The proposed network consists of a pose regressor and depth generator network. The former regresses 6 degree-of-freedom (DoF) pose values using CNN-RNN modules, and the latter generates depth maps using deep convolutional generative adversarial network (GAN) \cite{goodfellow2014generative}. The model takes a sequence of monocular images to estimate 6-DoF camera motion and depth map that is sampled from the same input data distribution, which is trained in an end-to-end and unsupervised fashion directly from raw input pixels. A view reconstruction approach is utilized as part of the objective function for the training in a similar way to prior works \cite{szeliski1999prediction, li2017undeepvo, zhou2017unsupervised, flynn2016deepstereo}. The entire pose estimation and depth map reconstruction pipeline is a consistent and systematic learning framework which continually improves its performance by collecting unlabelled monocular video data from numerous environments. This way, we want to mimic and transfer a continuous learning functionality from humans into VO domain.

In summary, the main contributions of our method are as follows: 
\begin{itemize}
	\item To the best of our knowledge, this is the first monocular VO method in literature, which uses adversarial and recurrent unsupervised learning approaches for joint pose and depth map estimation.
	\item We propose a novel adversarial technique for GANs to generate depth images without any need for depth groundtruth information.
	\item No strict parameter tuning is necessary for pose and depth estimation, contrary to traditional VO approaches.
\end{itemize}

Evaluations made on the KITTI \cite{menze2015object} and Cityscapes \cite{cordts2016cityscapes} datasets prove the success of our pose estimation and depth map reconstruction approach. As the outline of this paper, the previous work in VO is discussed in section \ref{sec:rel_work}. Section \ref{sec:overview} gives an overview of the proposed approach named GANVO. Section \ref{sec:method} describes the proposed unsupervised deep learning architecture and its mathematical background in detail. Section \ref{sec:results} shows our quantitative and qualitative results with comparisons to the existing VO methods. Finally, section \ref{sec:conclusion} concludes the study with some interesting future directions.

\section{Related Work}
\label{sec:rel_work}

Camera motion and depth map estimations are well studied problems in machine vision domain. Many traditional techniques have been proposed in the last decade with state-of-the-art results. However, existing traditional techniques require accurate image correspondence between consecutive frames, which is frequently violated in challenging environments with low texture, complex scene structure, and occlusions \cite{newcombe2011dtam, klein2007parallel, furukawa2010towards, turan2018sparse}. Deep learning approaches are adopted in several components of the traditional techniques to solve problems such as feature extraction and matching, pose estimation, and optical flow calculation. An external supervised training has played a key role in deep learning approaches to solve these aforementioned issues. 

Deep VO techniques involve 6-DoF camera pose estimation, depth map reconstruction, object segmentation, optical flow extraction, and sensor fusion approaches \cite{muller2017flowdometry, wang2017deepvo, bergen1992hierarchical, turan2017endosensorfusion, zhao2018learning, saputra2018visual, turan2017endo, turan2017deepfusion, turan2018magnetic}. Deep 6-DoF pose regression from raw RGB images was first proposed using CNNs, which was also extended to raw RGB-D images for challenging environments \cite{bo2013unsupervised}. Using RNNs that captures the temporal dynamics boosted the accuracy of 6-DoF pose estimation in deep VO approaches, resulting in a competitive performance against model-based VO methods \cite{wang2017deepvo}. The unsupervised deep learning studies on simultaneous estimation of the camera motion, image depth, surface normal and optical flow to learn structure from motion indicate that the joint training of VO components provides an implicit supervisory signal to the network \cite{godard2017unsupervised, zhou2017unsupervised, li2017undeepvo}. One critical issue of these unsupervised studies is the fact that they use auto encoder-decoder based traditional depth estimators with a tendency to generate overly smooth images \cite{dosovitskiy2016generating}. To solve this, we apply GANs that provide shaper and more accurate depth maps. The second issue of the aforementioned unsupervised techniques is the fact that they only employ CNNs that only analyse just-in-moment information to estimate camera pose \cite{wang2017deepvo, turan2018deep}. We address this issue by employing a CNN-RNN architecture to capture temporal relations across frames.   

\section{Architecture Overview}
\label{sec:overview}
As shown in Fig. \ref{fig:model_overview}, the raw RGB sequences consisting of a target view and source views are stacked together to form an input batch to the multi-view pose estimation module. The module regresses the relative 6-DoF pose values of the source views with respect to the target view. In parallel, the depth generation module generates a depth map of the target view. The view reconstruction module synthesizes the target image using the generated depth map, estimated 6-DoF camera pose and nearby colour values from source images. The view reconstruction constraint that provides a supervision signal forces the neural network to synthesize a target image from multiple source images acquired from different camera poses. The view discriminator module tries to distinguish the synthesized target image from the original target image. 

In the proposed adversarial scheme for a GAN, the objective of the generator is to trick the discriminator, i.e., to generate depth map for the target view reconstruction such that the discriminator cannot distinguish the reconstruction from the original. As opposed to the typical GANs, the output image of the generator is mapped to the color space of the image and the discriminator distinguishes this mapping from the original rather than a direct comparison of the generator output. The proposed scheme enables us to generate depth maps in an unsupervised manner.

\section{Unsupervised Depth and Pose Estimation with Generative Adversarial Network}
\label{sec:method}

\begin{figure*}
\includegraphics[width=\textwidth]{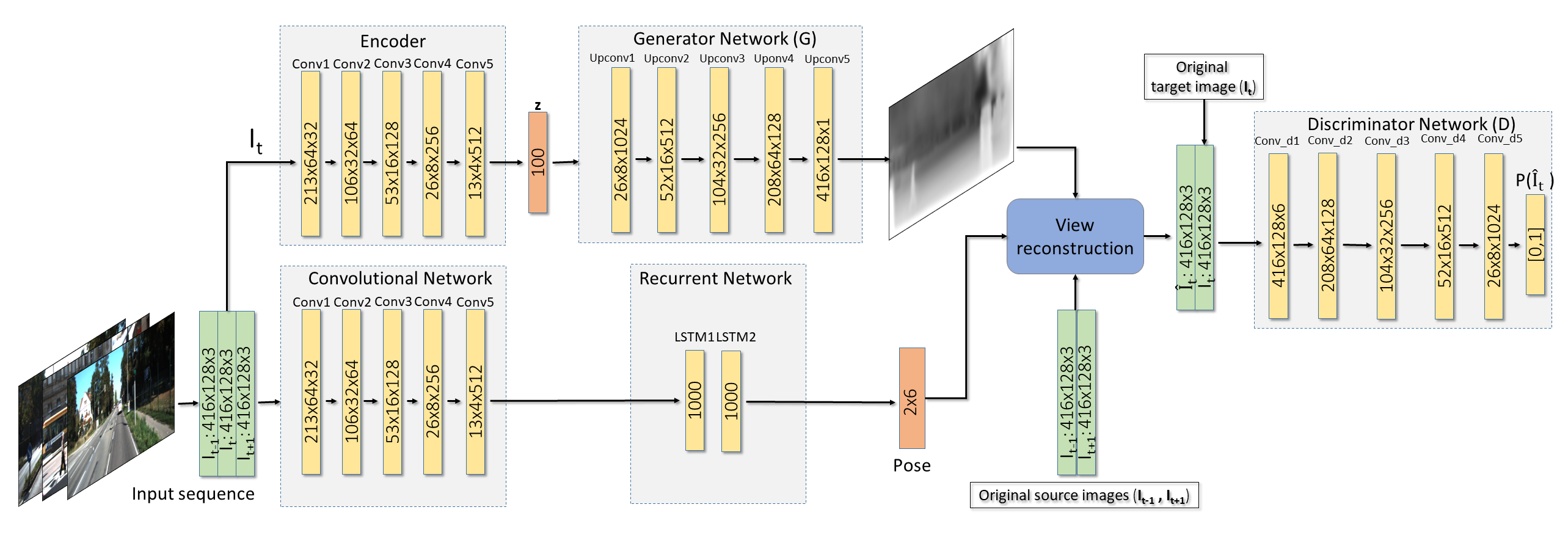}
\caption{The proposed architecture for pose estimation and depth map generation. The spatial dimensions of layers and output channels show the tensor shapes that flow through the network. Generator network $G$ maps a feature vector $\textbf{z}$ generated by the encoder network $E$ to the depth image space. The pose estimation network consists of a convolutional network that extracts VO related features, and a recurrent network that captures temporal relations among the input frame sequences. The pose results are collected after the recurrent network, which has $6*(N-1)$ output channels for 6-DoF motion parameters, , where $N$ is the length of the sequence. The view reconstruction algorithm is followed by a discriminator network $D$ that maps the reconstructed RGB image to a likelihood of the target image. $D$ decides whether it is shown the reconstruction or the original image.}
\label{fig:deeparchitecture}
\end{figure*}

The network architecture of the proposed method is shown in Fig. \ref{fig:deeparchitecture}. The details of the architecture are explained in the following sections.

\subsection{Depth Generator}
The first part of the architecture is a depth network that generates a single-view depth map of the target frame. The depth network is based on a GAN design that tries to learn the probability distribution of the input images $p(\textbf{I}_t)$, consisting of a generator network, $G$, a discriminator network $D$, and an encoder network $E$. The encoder $E$ maps the input target image $\textbf{I}_t$ to a feature vector $\textbf{z}$, i.e. $E(\textbf{I}_t) = \textbf{z}$. $G$ maps the feature vector $\textbf{z}$ to the depth image space and $D$ maps an input RGB image to an image likelihood.

\subsection{Pose Regressor}
The second network shown in the bottom of Fig. \ref{fig:deeparchitecture} tries to estimate relative pose $\textbf{p} \in\mathbb{SE}(3)$ introduced by motion and temporal dynamics across frames. The convolution part of the pose estimation network extracts features from the input frame sequences and propagates them to the RNN part. The LSTM modules output $6*(N-1)$ channels for 6-DoF pose values consisting translation and rotation parameters, where $N$ is the length of the sequence. Although one LSTM is able to capture the sequential relations, a second LSTM improves the learning the capacity of the network, resulting in a more accurate pose regression.

\subsection{View Reconstruction}

A sequence of $3$ consecutive frames is given to the pose network as input. An input sequence is denoted by $<I_{t-1}, I_{t}, I_{t+1}>$ where $t>0$ is the time index, $I_{t}$ is the target view, and the other frames are source views $I_{s} = <I_{t-1}, I_{t+1}>$ that are used to render the target image according to the objective function:
\begin{equation}
\mathcal{L}_{g} = \sum_{s}\sum_p | I_t (p) - \hat{I}_s(p) |
\end{equation}
where $p$ is the pixel coordinate index, and $\hat{I}_s$ is the projected image of the source view $I_s$ onto the target coordinate frame using a depth image based rendering module. The rendering is based on the estimated depth image $\hat{D}_t$, the $4\times 4$ camera transformation matrix $\hat{T}_{t\rightarrow s}$ and the source view $I_s$ \cite{fehn2004depth}. We denote the homogeneous coordinates of a pixel in the target view as $p_t$, and the camera intrinsics matrix as $K$. Coordinates of $p_t$ are projected onto the source view $p_s$ with:
\begin{equation}
p_s \sim K \hat{T}_{t\rightarrow s} \hat{D}_t(p_t) K^{-1} p_t.
\end{equation}

Since the value of $p_s$ is not discrete, an interpolation is required to find the expected intensity value at that position. To do that, we use bilinear interpolation using the $4$ discrete neighbours of $p_s$   \cite{zhou2016view}. The mean intensity value for projected pixel is estimated as follows:
\begin{equation}
\hat{I}_s(p_t) = I_s(p_s) = \sum_{i\in\{top,bottom\}, j\in\{left,right\}}w^{ij}I_s(p_s^{ij})
\end{equation}
where $w^{ij}$ is the proximity value between projected and neighbouring pixels, which sums up to $1$.

\subsection{View Discriminator}
 A realistic image is synthesized by the view reconstruction algorithm using the depth image generated by $G$ and estimated pose values. $D$ discriminates between the reconstructed image and the real image sampled from the target data distribution $p_{data}$, playing an adversarial role. These networks are trained by optimizing the objective loss function:
\begin{equation}
\begin{split}
\mathcal{L}_{d} = \min_{G} \max_{D} V(G,D) = &\mathbb{E}_{\textbf{I} \sim p_{data}(\textbf{I})} [\log(D(\textbf{I}))] + \\
&\mathbb{E}_{\textbf{z} \sim p(\textbf{z})} [\log(1-D(G(\textbf{z})))],
\end{split}
\end{equation}
where $\textbf{I}$ is the sample from the $p_{data}$ distribution and $\textbf{z}$ is a feature encoding on the latent space. 

\subsection{The Adversarial Training}
In contrast to the original GAN \cite{goodfellow2014generative}, we remove fully connected hidden layers for deeper architectures and use batchnorm in $G$ and $D$ networks. Pooling layers are replaced by strided convolutions and LeakyReLU activation is used for all layers in $D$. In $G$ network, pooling layers are replaced by fractional-strided convolutions and ReLU activation is used for all layers except for the output layer that uses tanh non-linearity. The GAN with these modifications generates non-blurry images and resolves the convergence problem during the training \cite{radford2015unsupervised}. The final objective for the optimization of weights in the architecture is:
\begin{equation}
\mathcal{L}_{final} = \mathcal{L}_{g} + \beta \mathcal{L}_{d}
\end{equation}
where $\beta$ is the balance factor. The optimal $\beta$ is experimentally found to be the ratio between the expected values $\mathcal{L}_{g}$ and $\mathcal{L}_{d}$ at the end of the training.

\section{Experiments and Results}
\label{sec:results}

We implemented the architecture with the publicly available Tensorflow framework \cite{abadi2016tensorflow}. Batch normalization is employed for all of the layers except for the output layers. The weights of the network are optimized with Adam optimization to increase the convergence rate, with the parameters $\beta_1 = 0.9$, $\beta_2 = 0.999$, learning rate of $0.1$ and mini-batch size of $8$. For training purposes, the input tensors of the model are assigned to sequential images of size $128 \times 416$, whereas they are not limited to any specific image size at test time. Three consecutive images are stacked together to form the input batch. We use the KITTI dataset \cite{geiger2012we} for benchmarking and the Cityscapes dataset \cite{cordts2016cityscapes} for evaluating cross-dataset generalization ability in the experiments. The model is trained on a NVIDIA TITAN V model GPU. We compare the proposed method with standard training/test splits on the KITTI dataset for the odometry and monocular depth map estimation tasks.

\subsection{Pose estimation benchmark}
\begin{figure}

	\begin{subfigure}{\columnwidth} 
	    \includegraphics[width=1.0\columnwidth]{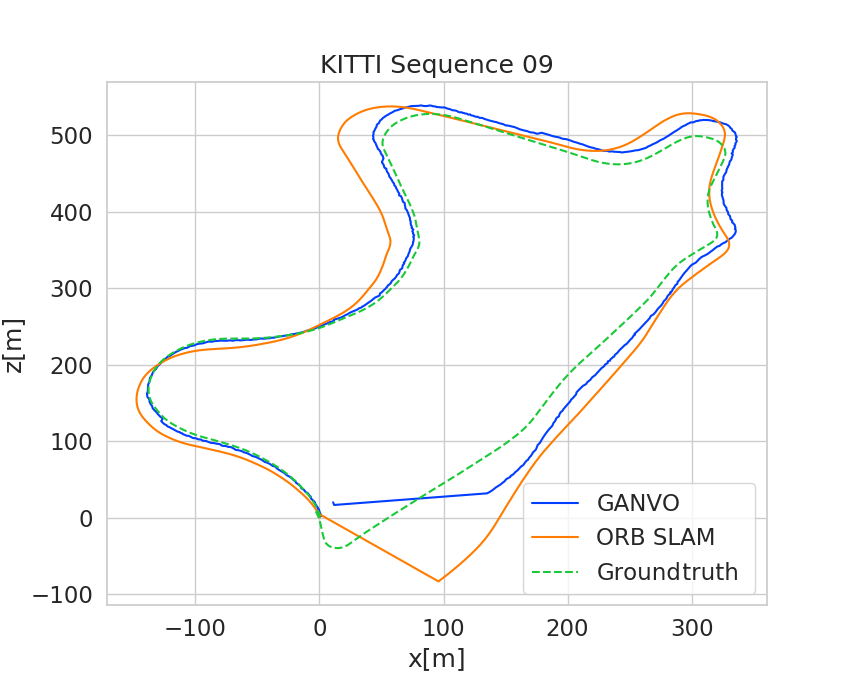}
        \label{fig:seq09}
	\end{subfigure}
	~
	\begin{subfigure}{\columnwidth} 
	    \includegraphics[width=1.0\columnwidth]{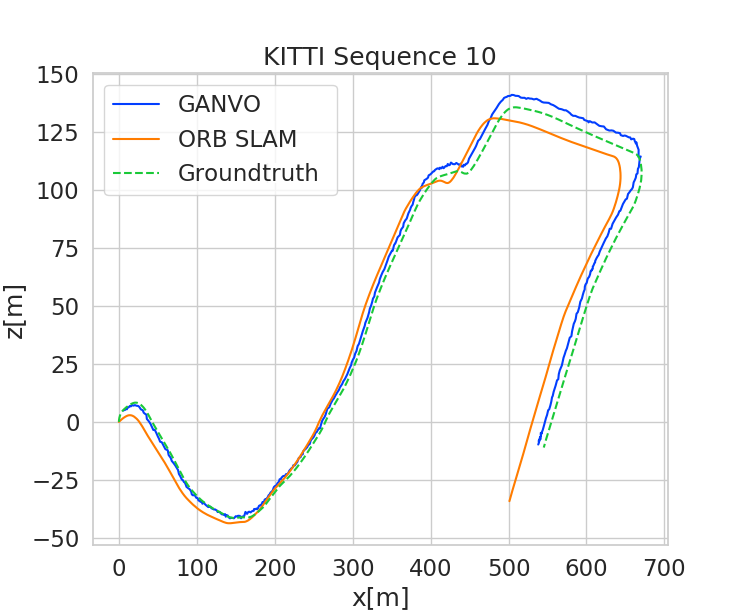}
        \label{fig:seq10}
	\end{subfigure}	
	\caption{Sample trajectories comparing the proposed unsupervised learning method GANVO with ORB SLAM, and the ground truth in meter scale. GANVO shows a better odometry estimation in terms of both rotational and translational motions. }
    \label{fig:traj_results}
\end{figure}

We have evaluated the pose estimation performance of our GANVO on the standard KITTI visual odometry split. The dataset contains $11$ driving sequences with groundtruth odometry obtained through the IMU/GPS sensors, where the sequences $00-08$ are used for training and $09-10$ for testing without any use of the pose and depth groundtruth during the training session. The network regresses the pose predictions as 6-DoF relative motion (Euclidean coordinates for translation and rotation) between sequences. We compare the pose estimation accuracy with the existing unsupervised deep learning approaches with the same sequence length of $5$, and monocular ORB SLAM. The results are evaluated using Absolute Trajectory Error (ATE) \cite{mur2015orb} for $5$ consecutive input frames with an optimized scaling factor to resolve scale ambiguity, which is reported to be the best sequence length for the compared methods \cite{zhou2017unsupervised, yin2018geonet}. As shown in Table \ref{tab:pose} and Fig. \ref{fig:traj_results}, the proposed method outperforms all the competing unsupervised and traditional baselines, without any need of global optimization steps such as loop closure detection, bundle adjustment and re-localization, revealing that GANVO captures long-term high level odometry details in addition to short-term low level odometry features.

\begin{table}[tbh]
\small
\begin{center}
\setlength{\tabcolsep}{5.0pt}
\begin{tabular*}{1.0\linewidth}{c|c|c}
\hline
Method & Seq.09 & Seq.10 \\
\hline
ORB-SLAM & $0.014\pm 0.008$ & $0.012\pm 0.011$\\
SfM-Learner \cite{zhou2017unsupervised} & $0.016\pm 0.009$ & $0.013\pm 0.009$\\
GeoNet \cite{yin2018geonet} &0.012 $\pm$ 0.007 & 0.012 $\pm$ 0.009\\
Our GANVO &\bf{ 0.009 $\pm$ 0.005 }& \bf{0.010 $\pm$ 0.013}\\
\hline
\end{tabular*}
\end{center}
\caption{Absolute Trajectory Error (ATE) on KITTI odometry dataset. We also report the results of the other methods for comparison that are taken from \cite{zhou2017unsupervised, yin2018geonet}. Our method outperforms all of the other methods.}
\label{tab:pose}
\vspace{-2ex}
\end{table}

\subsection{Single-view depth evaluation}

\begin{figure*}
\centering
\includegraphics[width=\textwidth]{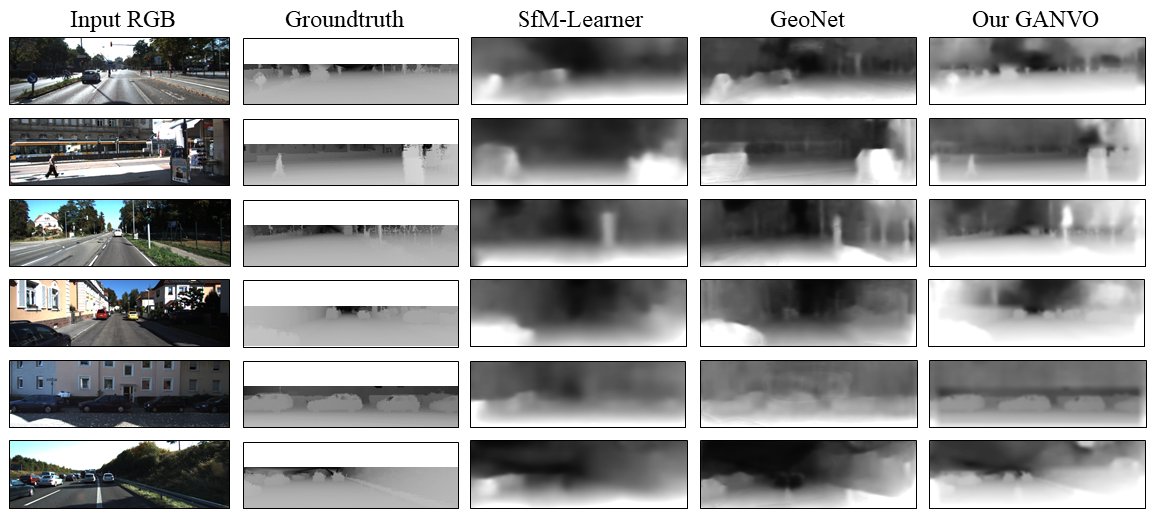}
\caption{ Comparison of unsupervised monocular depth estimation between SfM-Learner, GeoNet and the proposed GanVO. The groundtruth captured by Light Detection and Ranging (LiDAR) is interpolated for visualization purpose. GanVO captures details in challenging scenes containing low textured areas, shaded regions, and uneven road lines, preserving sharp, accurate and detailed depth map predictions both in close and distant regions.}
\label{fig:res_depth}
\end{figure*}

\begin{figure*}
\centering
\includegraphics[width=\textwidth]{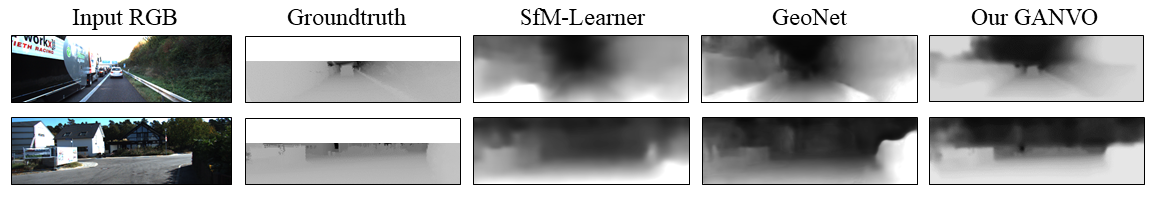}
\caption{ Typical failure cases of our model. Sometimes, all of the compared methods struggle in vast open rural scenes and huge objects occluding the camera view.}
\label{fig:res_depth_failure}
\end{figure*}

\begin{table*}[t]
\begin{center}
\small
\setlength{\tabcolsep}{5.0pt}
\begin{tabular*}{1.0\linewidth}{c|c|c|c|c|c|c|c|c|c}
\hline
Method & Supervision & Dataset & Abs Rel & Sq Rel & RMSE & RMSE log & $\delta<1.25$ & $\delta<1.25^2$ & $\delta<1.25^3$\\
\hline
Eigen \cite{eigen2014depth} Coarse & Depth & K & 0.214 & 1.605 & 6.563 & 0.292 & 0.673 & 0.884 & 0.957\\
Eigen \cite{eigen2014depth} Fine & Depth & K & 0.203 & 1.548 & 6.307 & 0.282 & 0.702 & 0.890 & 0.958 \\
Liu \cite{liu2016learning} & Depth & K & 0.202 & 1.614 & 6.523 & 0.275 & 0.678 & 0.895 & 0.965\\
Monodepth \cite{godard2017unsupervised}  & Pose & K & \bf{0.148} & 1.344 & 5.927 & 0.247 & 0.803 & 0.922 & 0.964\\ 
SfM-Learner \cite{zhou2017unsupervised} & No & K & 0.183 & 1.595 & 6.709 & 0.270 & 0.734 & 0.902 & 0.959\\
GeoNet \cite{yin2018geonet} & No & K & 0.155 & 1.296 & 5.857 & 0.233 & 0.793 & 0.931 & 0.973\\
GANVO & No & K & 0.150 & \bf{1.141} & \bf{5.448} & \bf{0.216} & \bf{0.808} & \bf{0.939} & \bf{0.975}\\
\hline
Garg et al. \cite{garg2016unsupervised} cap 50m & Pose & K & 0.169 & 1.080 & 5.104 & 0.273 & 0.740 & 0.904 & 0.962 \\
SfM-Learner \cite{zhou2017unsupervised} cap 50m & No & K & 0.201 & 1.391 & 5.181 & 0.264 & 0.696 & 0.900 & 0.966 \\
GeoNet \cite{yin2018geonet} cap 50m & No & K & 0.147 & 0.936 & 4.348 & 0.218 & 0.810 & 0.941 & 0.977\\
GANVO cap 50m & No & K & \bf{0.137} & \bf{0.892} & \bf{3.671} & \bf{0.201} & \bf{0.867} & \bf{0.970} & \bf{0.983}\\
\hline
Monodepth \cite{godard2017unsupervised}  & Pose & CS + K & \bf{0.124} & \bf{1.076} & \bf{5.311} & \bf{0.219} & \bf{0.847} & \bf{0.942} & 0.973\\
SfM-Learner \cite{zhou2017unsupervised} & No & CS + K & 0.198 & 1.836 & 6.565 & 0.275 & 0.718 & 0.901 & 0.960 \\
GeoNet \cite{yin2018geonet} & No & CS + K & 0.153 & 1.328 & 5.737 & 0.232 & 0.802 & 0.934 & 0.972\\
GANVO & No & CS + K & 0.138 & 1.155 & 4.412 & 0.232 & 0.820 & 0.939 & \bf{0.976}\\
\hline
\end{tabular*}
\end{center}
\caption{Monocular depth estimation results on the KITTI dataset \cite{menze2015object} by a benchmark split \cite{eigen2014depth}. The use of the KITTI dataset in the training is shown with the letter K, and the Cityscapes dataset \cite{cordts2016cityscapes} with CS. We report the error and accuracy values for the other methods for comparison taken from \cite{godard2017unsupervised, zhou2017unsupervised, yin2018geonet}. The best results are shown in bold. Garg et al. \cite{garg2016unsupervised} report 50m cap and we list them in a separate row for comparison.}
\label{tab:depth}
\vspace{-2ex}
\end{table*}

We evaluate the performance of the proposed depth estimation approach on a benchmark split of the KITTI dataset \cite{eigen2014depth} to compare with the existing learning-based and traditional depth estimation approaches. In a total of $44,000$ frames, we use $40,000$ frames for training and $4,000$ frames for validation. The sequence length of the input data is set to be $3$ frames during the training session to have the same evaluation setup with the compared methods, where the central frame is the target view for the depth estimation. The groundtruth captured by Light Detection and Ranging (LiDAR) sensor is projected into the image plane for the evaluation in terms of error and accuracy metrics. The predicted depth map, $D_{p}$, is multiplied by a scaling factor, $\hat{s}$, that matches the median with the groundtruth  depth map, $D_{g}$, to solve the scale ambiguity issue, i.e. $\hat{s} = median(D_g) / median(D_p)$. Moreover, we test the adaptability of the proposed approach training on the Cityscapes dataset and finetuning on the KITTI dataset.

Figure \ref{fig:res_depth} shows example depth map results predicted by the proposed method, SfM-Learner \cite{zhou2017unsupervised} and GeoNet \cite{yin2018geonet}. It is clearly seen that GANVO outputs sharper and more accurate depth maps compared to the other methods that fundamentally use an encoder-decoder network with various implementations. An explanation for this result is that adversarial training using the convolutional domain-related feature set of the discriminator allows to obtain less blurry results \cite{dosovitskiy2016generating}. Furthermore, it is also seen in Fig. \ref{fig:res_depth} that the depth maps predicted by the proposed GANVO  captures the small objects  in the scene whereas the other methods tend to ignore them. A loss function in image space leads to averaging all likely locations of details, whereas an adversarial loss function in feature space with a natural image prior makes the proposed GANVO more sensitive to the likely positions of the details in the scene \cite{dosovitskiy2016generating}. The proposed GANVO also accurately predicts the depth values of the objects in low-textured areas caused by the shading inconsistencies in a scene.  In Fig. \ref{fig:res_depth_failure}, we demonstrate typical failure cases in the compared unsupervised methods, which are caused by challenges such as poor road signs in rural areas and huge objects covering the most of the visual input. Even in these cases, GANVO performs slightly better than the existing methods.

As shown in Table \ref{tab:depth} quantitatively, our unsupervised approach significantly outperforms both the existing unsupervised methods \cite{garg2016unsupervised, zhou2017unsupervised, yin2018geonet} and even supervised methods \cite{eigen2014depth, liu2016learning}. Furthermore, in the benchmark where the approaches are compared in terms of their adaptability to different environments, the approaches are trained on the Cityscapes dataset and finetuned on the KITTI dataset. In this benchmark, the proposed method results in clearly better error and accuracy compared to the existing unsupervised methods. Moreover, GANVO obtains much closer results to Monodepth \cite{godard2017unsupervised} which is supervised by left-right image consistency, i.e. pose.

\begin{figure}
\centering
\includegraphics[width=\columnwidth]{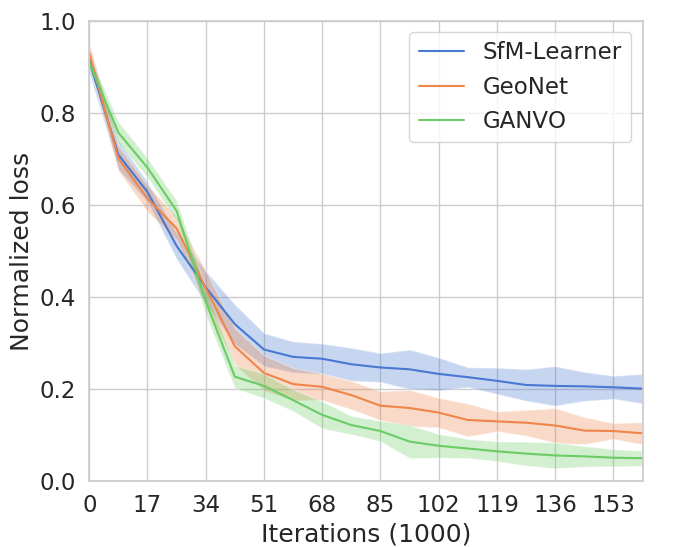}
\caption{Normalized loss values of GANVO compared to the unsupervised methods SfM-Learner \cite{zhou2017unsupervised} and GeoNet \cite{yin2018geonet} for multiple training experiments with various set of hyperparameters. GANVO is less sensitive to set of hyperparameters with lower mean and variance of the normalized loss values.}
\label{fig:training_loss}
\end{figure}  

Figure \ref{fig:training_loss} displays a robustness analysis of the proposed GANVO and the existing unsupervised approaches in terms of a parameter sensitivity analysis. We demonstrate that the proposed architecture is robust to hyper-parameter tunings such as different initialization of the network weights, different split of the dataset, and change of optimization parameters.

\section{CONCLUSIONS}
\label{sec:conclusion}
In this study, we proposed an unsupervised generative deep learning method for pose and depth map estimation for a monocular video sequences, demonstrating the effectiveness of adversarial learning in these tasks. The proposed method outperforms all the competing unsupervised and traditional baselines in terms of pose estimation, and captures more detailed, sharper and more accurate depth maps of the scenes. In a future work, we would like to explicitly address scene dynamics and investigate a learning representation for a full 3D volumetric modelling with semantic segmentation.

\paragraph*{Acknowledgments} We would like to thank the authors referred in this study for sharing their code. This work is funded by the NIST grant 70NANB17H185. YA would like to thank the Ministry of National Education in Turkey for their funding and support.

\bibliographystyle{IEEEtran}
\balance


\begin{thebibliography}{10}
\providecommand{\url}[1]{#1}
\csname url@samestyle\endcsname
\providecommand{\newblock}{\relax}
\providecommand{\bibinfo}[2]{#2}
\providecommand{\BIBentrySTDinterwordspacing}{\spaceskip=0pt\relax}
\providecommand{\BIBentryALTinterwordstretchfactor}{4}
\providecommand{\BIBentryALTinterwordspacing}{\spaceskip=\fontdimen2\font plus
\BIBentryALTinterwordstretchfactor\fontdimen3\font minus
  \fontdimen4\font\relax}
\providecommand{\BIBforeignlanguage}[2]{{%
\expandafter\ifx\csname l@#1\endcsname\relax
\typeout{** WARNING: IEEEtran.bst: No hyphenation pattern has been}%
\typeout{** loaded for the language `#1'. Using the pattern for}%
\typeout{** the default language instead.}%
\else
\language=\csname l@#1\endcsname
\fi
#2}}
\providecommand{\BIBdecl}{\relax}
\BIBdecl

\bibitem{menze2015object}
M.~Menze and A.~Geiger, ``Object scene flow for autonomous vehicles,'' in
  \emph{Proceedings of the IEEE Conference on Computer Vision and Pattern
  Recognition}, 2015, pp. 3061--3070.

\bibitem{cordts2016cityscapes}
M.~Cordts, M.~Omran, S.~Ramos, T.~Rehfeld, M.~Enzweiler, R.~Benenson,
  U.~Franke, S.~Roth, and B.~Schiele, ``The cityscapes dataset for semantic
  urban scene understanding,'' in \emph{Proceedings of the IEEE conference on
  computer vision and pattern recognition}, 2016, pp. 3213--3223.

\bibitem{engel2014lsd}
J.~Engel, T.~Sch{\"o}ps, and D.~Cremers, ``Lsd-slam: Large-scale direct
  monocular slam,'' in \emph{European Conference on Computer Vision}.\hskip 1em
  plus 0.5em minus 0.4em\relax Springer, 2014, pp. 834--849.

\bibitem{mur2017orb}
R.~Mur-Artal and J.~D. Tard{\'o}s, ``Orb-slam2: An open-source slam system for
  monocular, stereo, and rgb-d cameras,'' \emph{IEEE Transactions on Robotics},
  vol.~33, no.~5, pp. 1255--1262, 2017.

\bibitem{wang2017deepvo}
S.~Wang, R.~Clark, H.~Wen, and N.~Trigoni, ``Deepvo: Towards end-to-end visual
  odometry with deep recurrent convolutional neural networks,'' in
  \emph{Robotics and Automation (ICRA), 2017 IEEE International Conference
  on}.\hskip 1em plus 0.5em minus 0.4em\relax IEEE, 2017, pp. 2043--2050.

\bibitem{clark2017vinet}
R.~Clark, S.~Wang, H.~Wen, A.~Markham, and N.~Trigoni, ``Vinet: Visual-inertial
  odometry as a sequence-to-sequence learning problem.'' in \emph{AAAI}, 2017,
  pp. 3995--4001.

\bibitem{turan2018deep}
M.~Turan, Y.~Almalioglu, H.~Araujo, E.~Konukoglu, and M.~Sitti, ``Deep endovo:
  A recurrent convolutional neural network (rcnn) based visual odometry
  approach for endoscopic capsule robots,'' \emph{Neurocomputing}, vol. 275,
  pp. 1861--1870, 2018.

\bibitem{turan2017deep}
M.~Turan, Y.~Almalioglu, E.~Konukoglu, and M.~Sitti, ``A deep learning based 6
  degree-of-freedom localization method for endoscopic capsule robots,''
  \emph{arXiv preprint arXiv:1705.05435}, 2017.

\bibitem{muller2017flowdometry}
P.~Muller and A.~Savakis, ``Flowdometry: An optical flow and deep learning
  based approach to visual odometry,'' in \emph{Applications of Computer Vision
  (WACV), 2017 IEEE Winter Conference on}.\hskip 1em plus 0.5em minus
  0.4em\relax IEEE, 2017, pp. 624--631.

\bibitem{artetxe2017unsupervised}
M.~Artetxe, G.~Labaka, E.~Agirre, and K.~Cho, ``Unsupervised neural machine
  translation,'' \emph{arXiv preprint arXiv:1710.11041}, 2017.

\bibitem{turan2018unsupervised}
M.~Turan, E.~P. Ornek, N.~Ibrahimli, C.~Giracoglu, Y.~Almalioglu, M.~F. Yanik,
  and M.~Sitti, ``Unsupervised odometry and depth learning for endoscopic
  capsule robots,'' \emph{arXiv preprint arXiv:1803.01047}, 2018.

\bibitem{brown2016prospective}
T.~I. Brown, V.~A. Carr, K.~F. LaRocque, S.~E. Favila, A.~M. Gordon, B.~Bowles,
  J.~N. Bailenson, and A.~D. Wagner, ``Prospective representation of
  navigational goals in the human hippocampus,'' \emph{Science}, vol. 352, no.
  6291, pp. 1323--1326, 2016.

\bibitem{wamsley2010brief}
E.~J. Wamsley, M.~A. Tucker, J.~D. Payne, and R.~Stickgold, ``A brief nap is
  beneficial for human route-learning: The role of navigation experience and
  eeg spectral power,'' \emph{Learning \& Memory}, vol.~17, no.~7, pp.
  332--336, 2010.

\bibitem{goodfellow2014generative}
I.~Goodfellow, J.~Pouget-Abadie, M.~Mirza, B.~Xu, D.~Warde-Farley, S.~Ozair,
  A.~Courville, and Y.~Bengio, ``Generative adversarial nets,'' in
  \emph{Advances in neural information processing systems}, 2014, pp.
  2672--2680.

\bibitem{szeliski1999prediction}
R.~Szeliski, ``Prediction error as a quality metric for motion and stereo,'' in
  \emph{Computer Vision, 1999. The Proceedings of the Seventh IEEE
  International Conference on}, vol.~2.\hskip 1em plus 0.5em minus 0.4em\relax
  IEEE, 1999, pp. 781--788.

\bibitem{li2017undeepvo}
R.~Li, S.~Wang, Z.~Long, and D.~Gu, ``Undeepvo: Monocular visual odometry
  through unsupervised deep learning,'' \emph{arXiv preprint arXiv:1709.06841},
  2017.

\bibitem{zhou2017unsupervised}
T.~Zhou, M.~Brown, N.~Snavely, and D.~G. Lowe, ``Unsupervised learning of depth
  and ego-motion from video,'' in \emph{CVPR}, vol.~2, no.~6, 2017, p.~7.

\bibitem{flynn2016deepstereo}
J.~Flynn, I.~Neulander, J.~Philbin, and N.~Snavely, ``Deepstereo: Learning to
  predict new views from the world's imagery,'' in \emph{Proceedings of the
  IEEE Conference on Computer Vision and Pattern Recognition}, 2016, pp.
  5515--5524.

\bibitem{newcombe2011dtam}
R.~A. Newcombe, S.~J. Lovegrove, and A.~J. Davison, ``Dtam: Dense tracking and
  mapping in real-time,'' in \emph{Computer Vision (ICCV), 2011 IEEE
  International Conference on}.\hskip 1em plus 0.5em minus 0.4em\relax IEEE,
  2011, pp. 2320--2327.

\bibitem{klein2007parallel}
G.~Klein and D.~Murray, ``Parallel tracking and mapping for small ar
  workspaces,'' in \emph{Mixed and Augmented Reality, 2007. ISMAR 2007. 6th
  IEEE and ACM International Symposium on}.\hskip 1em plus 0.5em minus
  0.4em\relax IEEE, 2007, pp. 225--234.

\bibitem{furukawa2010towards}
Y.~Furukawa, B.~Curless, S.~M. Seitz, and R.~Szeliski, ``Towards internet-scale
  multi-view stereo,'' in \emph{Computer Vision and Pattern Recognition (CVPR),
  2010 IEEE Conference on}.\hskip 1em plus 0.5em minus 0.4em\relax IEEE, 2010,
  pp. 1434--1441.

\bibitem{turan2018sparse}
M.~Turan, Y.~Y. Pilavci, I.~Ganiyusufoglu, H.~Araujo, E.~Konukoglu, and
  M.~Sitti, ``Sparse-then-dense alignment-based 3d map reconstruction method
  for endoscopic capsule robots,'' \emph{Machine Vision and Applications},
  vol.~29, no.~2, pp. 345--359, 2018.

\bibitem{bergen1992hierarchical}
J.~R. Bergen, P.~Anandan, K.~J. Hanna, and R.~Hingorani, ``Hierarchical
  model-based motion estimation,'' in \emph{European conference on computer
  vision}.\hskip 1em plus 0.5em minus 0.4em\relax Springer, 1992, pp. 237--252.

\bibitem{turan2017endosensorfusion}
M.~Turan, Y.~Almalioglu, H.~Araujo, T.~Cemgil, and M.~Sitti,
  ``Endosensorfusion: Particle filtering-based multi-sensory data fusion with
  switching state-space model for endoscopic capsule robots using recurrent
  neural network kinematics,'' \emph{arXiv preprint arXiv:1709.03401}, 2017.

\bibitem{zhao2018learning}
C.~Zhao, L.~Sun, P.~Purkait, T.~Duckett, and R.~Stolkin, ``Learning monocular
  visual odometry with dense 3d mapping from dense 3d flow,'' \emph{arXiv
  preprint arXiv:1803.02286}, 2018.

\bibitem{saputra2018visual}
M.~R.~U. Saputra, A.~Markham, and N.~Trigoni, ``Visual slam and structure from
  motion in dynamic environments: A survey,'' \emph{ACM Computing Surveys
  (CSUR)}, vol.~51, no.~2, p.~37, 2018.

\bibitem{turan2017endo}
M.~Turan, Y.~Almalioglu, H.~Gilbert, A.~E. Sari, U.~Soylu, and M.~Sitti,
  ``Endo-vmfusenet: deep visual-magnetic sensor fusion approach for
  uncalibrated, unsynchronized and asymmetric endoscopic capsule robot
  localization data,'' \emph{arXiv preprint arXiv:1709.06041}, 2017.

\bibitem{turan2017deepfusion}
M.~Turan, J.~Shabbir, H.~Araujo, E.~Konukoglu, and M.~Sitti, ``A deep learning
  based fusion of rgb camera information and magnetic localization information
  for endoscopic capsule robots,'' \emph{International journal of intelligent
  robotics and applications}, vol.~1, no.~4, pp. 442--450, 2017.

\bibitem{turan2018magnetic}
M.~Turan, Y.~Almalioglu, E.~P. Ornek, H.~Araujo, M.~F. Yanik, and M.~Sitti,
  ``Magnetic-visual sensor fusion-based dense 3d reconstruction and
  localization for endoscopic capsule robots,'' \emph{arXiv preprint
  arXiv:1803.01048}, 2018.

\bibitem{bo2013unsupervised}
L.~Bo, X.~Ren, and D.~Fox, ``Unsupervised feature learning for rgb-d based
  object recognition,'' in \emph{Experimental Robotics}.\hskip 1em plus 0.5em
  minus 0.4em\relax Springer, 2013, pp. 387--402.

\bibitem{godard2017unsupervised}
C.~Godard, O.~Mac~Aodha, and G.~J. Brostow, ``Unsupervised monocular depth
  estimation with left-right consistency,'' in \emph{CVPR}, vol.~2, no.~6,
  2017, p.~7.

\bibitem{dosovitskiy2016generating}
A.~Dosovitskiy and T.~Brox, ``Generating images with perceptual similarity
  metrics based on deep networks,'' in \emph{Advances in Neural Information
  Processing Systems}, 2016, pp. 658--666.

\bibitem{fehn2004depth}
C.~Fehn, ``Depth-image-based rendering (dibr), compression, and transmission
  for a new approach on 3d-tv,'' in \emph{Stereoscopic Displays and Virtual
  Reality Systems XI}, vol. 5291.\hskip 1em plus 0.5em minus 0.4em\relax
  International Society for Optics and Photonics, 2004, pp. 93--105.

\bibitem{zhou2016view}
T.~Zhou, S.~Tulsiani, W.~Sun, J.~Malik, and A.~A. Efros, ``View synthesis by
  appearance flow,'' in \emph{European conference on computer vision}.\hskip
  1em plus 0.5em minus 0.4em\relax Springer, 2016, pp. 286--301.

\bibitem{radford2015unsupervised}
A.~Radford, L.~Metz, and S.~Chintala, ``Unsupervised representation learning
  with deep convolutional generative adversarial networks,'' \emph{arXiv
  preprint arXiv:1511.06434}, 2015.

\bibitem{abadi2016tensorflow}
M.~Abadi, P.~Barham, J.~Chen, Z.~Chen, A.~Davis, J.~Dean, M.~Devin,
  S.~Ghemawat, G.~Irving, M.~Isard \emph{et~al.}, ``Tensorflow: a system for
  large-scale machine learning.'' in \emph{OSDI}, vol.~16, 2016, pp. 265--283.

\bibitem{geiger2012we}
A.~Geiger, P.~Lenz, and R.~Urtasun, ``Are we ready for autonomous driving? the
  kitti vision benchmark suite,'' in \emph{Computer Vision and Pattern
  Recognition (CVPR), 2012 IEEE Conference on}.\hskip 1em plus 0.5em minus
  0.4em\relax IEEE, 2012, pp. 3354--3361.

\bibitem{mur2015orb}
R.~Mur-Artal, J.~M.~M. Montiel, and J.~D. Tardos, ``Orb-slam: a versatile and
  accurate monocular slam system,'' \emph{IEEE Transactions on Robotics},
  vol.~31, no.~5, pp. 1147--1163, 2015.

\bibitem{yin2018geonet}
Z.~Yin and J.~Shi, ``Geonet: Unsupervised learning of dense depth, optical flow
  and camera pose,'' in \emph{Proceedings of the IEEE Conference on Computer
  Vision and Pattern Recognition (CVPR)}, vol.~2, 2018.

\bibitem{eigen2014depth}
D.~Eigen, C.~Puhrsch, and R.~Fergus, ``Depth map prediction from a single image
  using a multi-scale deep network,'' in \emph{Advances in neural information
  processing systems}, 2014, pp. 2366--2374.

\bibitem{liu2016learning}
F.~Liu, C.~Shen, G.~Lin, and I.~D. Reid, ``Learning depth from single monocular
  images using deep convolutional neural fields.'' \emph{IEEE Trans. Pattern
  Anal. Mach. Intell.}, vol.~38, no.~10, pp. 2024--2039, 2016.

\bibitem{garg2016unsupervised}
R.~Garg, V.~K. BG, G.~Carneiro, and I.~Reid, ``Unsupervised cnn for single view
  depth estimation: Geometry to the rescue,'' in \emph{European Conference on
  Computer Vision}.\hskip 1em plus 0.5em minus 0.4em\relax Springer, 2016, pp.
  740--756.

\end{thebibliography}

\end{document}